\documentclass[lettersize,journal]{IEEEtran}
\usepackage{amsmath,amsfonts}
\usepackage{algorithmic}
\usepackage{algorithm}
\usepackage{array}
\usepackage[caption=false,font=normalsize,labelfont=sf,textfont=sf]{subfig}
\usepackage{textcomp}
\usepackage{stfloats}
\usepackage{url}
\usepackage{verbatim}
\usepackage{graphicx}
\usepackage{cite}
\usepackage{ multirow}
\usepackage{amssymb}
\usepackage{color,xcolor}
\begin{document}

\title{Weakly-Supervised Video Anomaly Detection with Snippet Anomalous Attention}

\author{Yidan Fan, Yongxin Yu, Wenhuan Lu, Yahong Han
\thanks{This work was supported by the CAAI-Huawei MindSpore Open Fund. Yidan Fan, Yongxin Yu, Wenhuan Lu and Yahong Han are with the College of Intelligence and Computing, Tianjin University, Tianjin 300072, China. E-mail: \{yidan\_fan, yyx, wenhuan, yahong\}@tju.edu.cn.}}

\markboth{Journal of \LaTeX\ Class Files,~Vol.~14, No.~8, August~2021}%
{Shell \MakeLowercase{\textit{et al.}}: A Sample Article Using IEEEtran.cls for IEEE Journals}


\maketitle

\begin{abstract}
With a focus on abnormal events contained within untrimmed videos, there is increasing interest among researchers in video anomaly detection. Among different video anomaly detection scenarios, weakly-supervised video anomaly detection poses a significant challenge as it lacks frame-wise labels during the training stage, only relying on video-level labels as coarse supervision. Previous methods have made attempts to either learn discriminative features in an end-to-end manner or employ a two-stage self-training strategy to generate snippet-level pseudo labels. However, both approaches have certain limitations. The former tends to overlook informative features at the snippet level, while the latter can be susceptible to noises. In this paper, we propose an Anomalous Attention mechanism for weakly-supervised anomaly detection to tackle the aforementioned problems. Our approach takes into account snippet-level encoded features without the supervision of pseudo labels. Specifically, our approach first generates snippet-level anomalous attention and then feeds it together with original anomaly scores into a Multi-branch Supervision Module. The module learns different areas of the video, including areas that are challenging to detect, and also assists the attention optimization. Experiments on benchmark datasets XD-Violence and UCF-Crime verify the effectiveness of our method. Besides, thanks to the proposed snippet-level attention, we obtain a more precise anomaly localization.
\end{abstract}

\begin{IEEEkeywords}
Weakly-supervised, Anomaly detection, Snippet anomalous attention, Multi-branch supervision.
\end{IEEEkeywords}

\section{Introduction}
Video anomaly detection (VAD) is a crucial task in the analysis of activities in untrimmed videos, aiming to detect unusual events within video frames or snippets.

Unsupervised VAD has gained significant attention from researchers due to its ability to detect anomalies without requiring additional annotations \cite{lai2021anomaly,zaheer2022generative,liu2021hybrid,ristea2022self,lu2022learnable,zhong2022bidirectional,zhang2020normality,zhou2019attention}. However, these approaches only have access to normal videos during the training phase, leading to a limited understanding of anomaly data. Consequently, unsupervised-VAD methods often exhibit a high false alarm rate for previously unseen normal events. In order to address the incorrect recognition of video anomalies in the unsupervised setting, a more practical scenario is considered where only the video level labels are available, i.e., weakly labeled abnormal or normal training videos. Addressing this scenario, the paper \cite{sultani2018real} first proposed weakly-supervised anomaly detection (WS-VAD). Compared to the unsupervised pipeline, the WS-VAD paradigm provides a better trade-off between detection performance and manual annotation cost.
\begin{figure}[t]
	\centering
	\includegraphics[width=\linewidth]{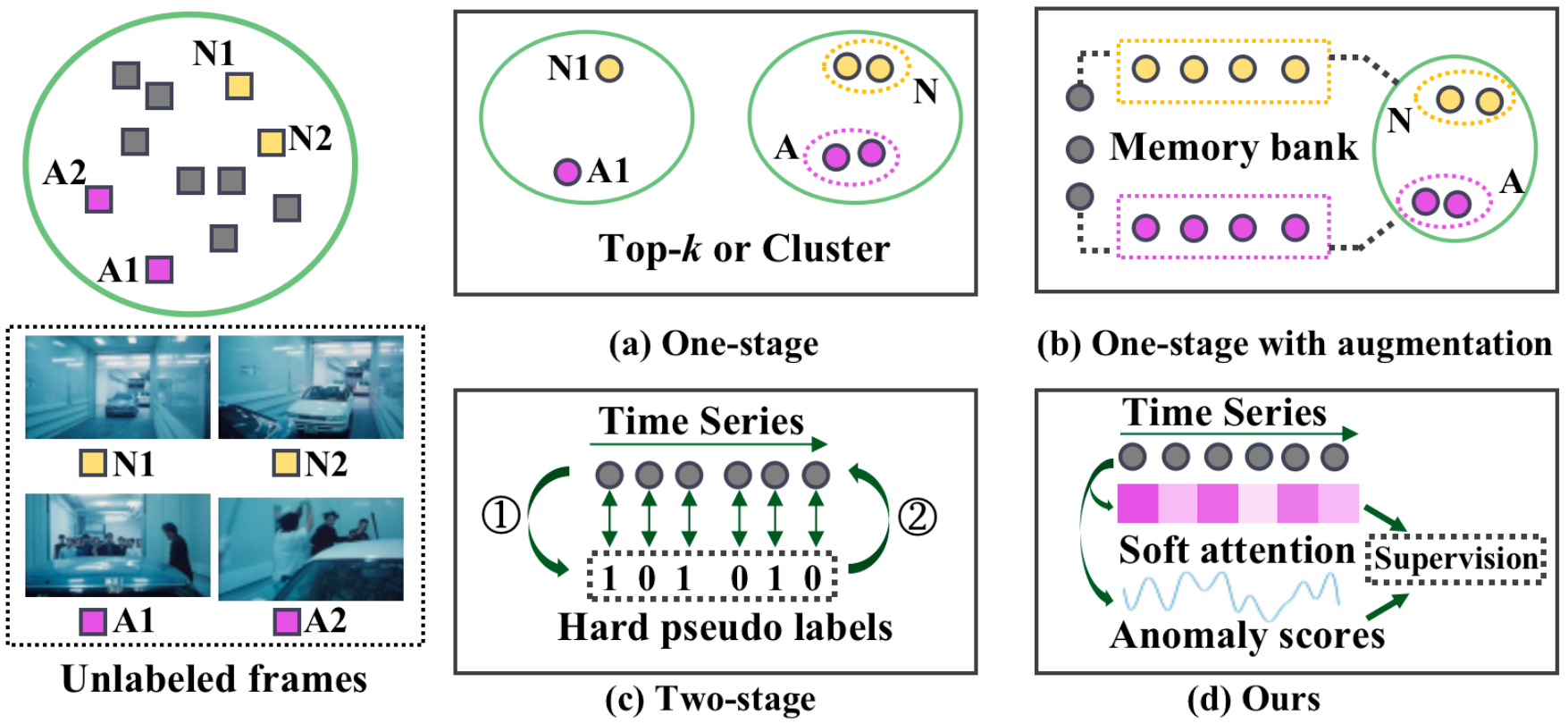}
	\caption{Comparisons with the existing approaches. \textit{N} refers to normal frames while \textit{A} refers to abnormal. Frames ``$\blacksquare$'' are firstly processed to snippet-level features ``$\bullet$''. In the one-stage method (a), normal and abnormal features are directly clustered or chosen with feature magnitude. In (b), additional memory banks are used to augment original features. While in (c), hard pseudo labels 0 or 1 are generated first and then used to direct snippet-level supervision. Labels are refined through the second stage. In our approach (d), snippet-level anomalous soft attention is generated first, and with general prediction scores, they are then fed into a multi-branch supervision module.}
	\label{fig:Comparison}
\end{figure}

In the field of WS-VAD, a multitude of methods have been proposed to fully utilization of these weak annotations and existing approaches can be broadly categorized into two types based on the steps employed to generate final abnormal predictions: one-stage methods based on Multiple Instance Learning (MIL)  \cite{sultani2018real,wu2021learning,wu2021weakly,purwanto2021dance,zhu2019motion,zhang2019temporal,tian2021weakly,zhou2023dual,chen2022mgfn,park2023normality}, and two-stage self-training methods \cite{feng2021mist,wu2022self,zhang2022exploiting,li2022self}. In the case of one-stage methods,
the key idea is to select representative abnormal and normal features, scores of these snippets are then used for final video-level classification. As shown in (a) of Figure \ref{fig:Comparison}, the top-$k$ selection based on feature magnitude is applied in \cite{tian2021weakly,chen2022mgfn} and in \cite{zaheer2020claws} a clustering distance-based loss is proposed to produce better anomaly representations. In \cite{park2023normality,zhou2023dual}, additional memory modules are used to augment the original feature for the purpose of learning discriminative features, shown in (b). As for the two-stage approaches, in \cite{feng2021mist,zhang2022exploiting}, snippet-level pseudo labels are generated in stage 1 and then refined through the backpropagation process in step 2, which is shown in (c).
\begin{figure*}[t]
	\centering
	\includegraphics[width=0.92\linewidth]{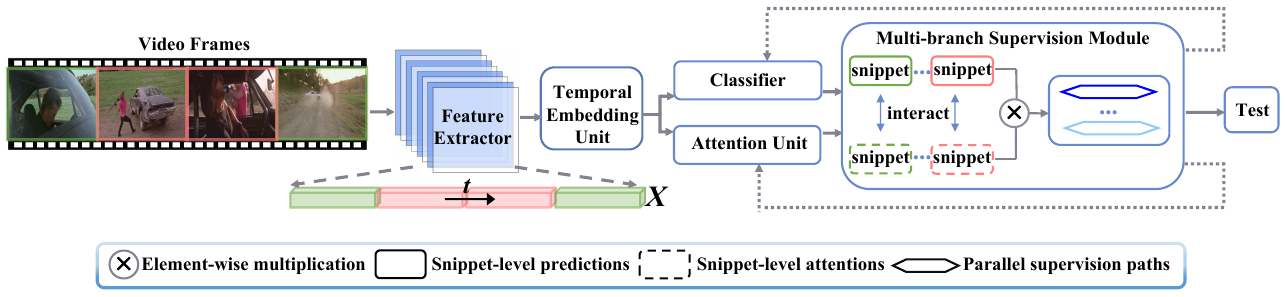}
	\caption{The proposed method consists of three primary modules: Temporal Embedding Unit, Anomalous Attention Unit, and Multi-branch Supervision Module. The first module is responsible for encoding the feature, while the second module focuses on detecting snippet-level anomalies and generating attention. The third module aims to model the completeness of anomaly. To generate anomalous attention, an optimization process (dash line in the figure) is designed.}
	\label{fig:structure}
\end{figure*}

Although previous one-stage methods have achieved good performance, they still have limitations in snippet-level feature understanding. These methods tend to focus only on representative snippets, resulting in informative features being overlooked during the selection process which is biased and can then cause normal segments surrounding anomalies to be highly misclassified. Besides if the selected instance is wrong in the initial training stage, errors will accumulate and lead to poor predictions. Moreover, introducing a memory module and updating it is inevitably time and resource-consuming. The two-stage methods try to tackle this problem with generated snippet-level pseudo labels. But pseudo labels are strong hints for supervision and contain much noise, thus also leading to unsatisfactory performance.

To address the aforementioned issues, our method first incorporates a temporal embedding unit to model the whole video, which aggregates both local and global information. Also, we adopt a soft attention mechanism to handle the WS-VAD task. Specifically, anomalous attention in the temporal dimension is generated to fully utilize intermediate snippet-level embeddings and guide the supervision process in a soft manner. Moreover, some anomalies are difficult to be distinguished due to their slight difference from normal events or only occupy a small portion of the frames, prompting us to propose a multi-branch supervision module, i.e., general supervision, attention-based supervision, general suppressed and attention-based suppressed supervision to explore the completeness of anomaly and detect difficult anomalous areas. As a result, a more robust anomaly-detected model can be obtained.

Previous methods for WS-VAD have incorporated attention mechanisms such as \cite{zhu2019motion} and \cite{zhang2019temporal}, but our approach is different from them in several ways. Specifically, both methods add attention branches with structures similar to the primary classifier branch while our attention unit is completely different from the classifier branch. In \cite{zhu2019motion}, the output of the attention branch tries to capture the total anomaly score of the entire video, while in \cite{zhang2019temporal}, anomalous scores from both branches are simply averaged and used for final video classification. Unlike these methods, our attention mechanism is category-agnostic and anomaly-specific.  Moreover, our attention measures anomalies in each snippet and is optimized by anomalous scores, rather than video-level labels.

In summary, the main contributions of this paper are as follows:
\begin{itemize}
	\item We introduce a snippet-level attention mechanism using the intermediate embeddings from the consideration that they contain more semantic information and are beneficial for final frame-level \footnote{16 frames per snippet in our method like others.} anomaly detection tasks. The attention is anomaly-specific and not optimized by video-level annotations but rather by anomaly predictions.
	\item With the assistance of anomalous attention in a soft manner, we propose a multi-branch supervision module to better explore the hard abnormal part of the whole video. Also, the completeness of the anomaly events and accuracy of the localization can be achieved.
	\item To validate our approach, we conducted experiments on two benchmark datasets: UCF-Crime and XD-Violence, and the state-of-the-art performance verify the effectiveness of our method.
\end{itemize}

\section{Related Work}
\subsection{Weakly-supervised Anomaly Detection}
With limited labels, the objective of WS-VAD is to generate final frame-level anomalous scores. As mentioned before, previous approaches to WS-VAD can be broadly classified into two types: one-stage methods based on Multiple Instance Learning (MIL), and two-stage self-training methods.

Regarding one-stage methods, \cite{sultani2018real} is the first study to introduce the MIL frameworks to WS-VAD. Besides, in \cite{sultani2018real}, hinge loss is employed, and anomaly scores of abnormal instances are enforced to be greater than the normal ones. Later, \cite{tian2021weakly} notes that WS-VAD is biased by the dominant negative instances, especially when the abnormal events have subtle differences from normal events. Then they train a feature magnitude learning function to effectively recognize the positive instances. Then in \cite{chen2022mgfn}, the author points out that feature magnitudes to represent the degree of anomalies typically ignore the effects of scene variations and thus propose a feature amplification mechanism and a magnitude contrastive loss to enhance the discriminativeness of features for detecting anomalies. Similar to the unsupervised anomaly detection methods, memory modules that can store the representative patterns are introduced in \cite{park2023normality,zhou2023dual}. The former encodes diverse normal features into prototypes and then constructs a similarity-based classifier. The latter uses two memory banks, one to store representative abnormal patterns and the other to store normal ones.

In the case of two-stage self-training methods, \cite{feng2021mist} introduced a multiple instance pseudo-label generator and a self-guided attention-boosted feature encoder to refine task-specific representations. In \cite{li2022self}, a self-training strategy that gradually refines anomaly scores is proposed based on Multi-Sequence Learning (MSL). Furthermore, \cite{zhang2022exploiting} presents a multi-head classification module and an iterative uncertainty pseudo-label refinement strategy.

Our approach to anomaly modeling differs from previous methods in two ways. Firstly, although our method also follows the one-stage MIL pipeline, it does not aim to choose representative features. Instead, the most discriminative part of the untrimmed video is suppressed in our approach. Secondly, from the perspective of modeling the anomaly completeness, although our approach has similar motivation with \cite{zhang2022exploiting} which introduces a multi-head classification module,  our multi-branch supervision module only utilizes one classifier and obtains diverse anomalous score sequences based on attention, thus effectively exploring anomaly completeness.
\subsection{Weakly-supervised Temporal Action Localization}
Weakly supervised temporal action localization is an efficient method for understanding human action \cite{tu2018ml} instances without overwhelming annotations. Several works \cite{paul2018w, islam2020weakly} have addressed this problem also using the multiple-instance learning (MIL) framework and primarily rely on aggregating class scores at the snippet level to generate video-level predictions, which is similar to the strategy generally use in the WS-VAD field. While others like \cite{islam2021hybrid,luo2020weakly,qu2021acm,hong2021cross,zhao2023novel,lee2020background}  treat the background frames as an auxiliary class. Then utilizing the complementary learning scheme or filtering irrelevant information scheme at the snippet level to ensure precise positioning accuracy. Our work is mainly inspired by this snippet-level focus broadly used in WS-TAL tasks, but different from them in several ways. Firstly, any video in WS-TAL comprises both action frames and background frames, with background events typically regarded as an auxiliary class. Conversely, for WS-TAD, the normal subset only includes normal events. Secondly, WS-TAL is a multi-label classification task, the final result is the probability of each category, including the background class. On the other hand, WS-TAD is a regression task that generates precise anomaly scores as the final outcome. Lastly, in WS-TAL, the number of categories is known, and the given labels are exact action categories occurring in all videos, while for WS-TAD, the anomalies are varied and class-unknown. Therefore, despite our work drawing inspiration from WS-TAL, action temporal localization, and anomaly detection are significantly different tasks.
\section{Methods}
In this paper, we propose a Snippet-level Anomalous Attention-based Multi-branch Supervision framework for WS-VAD task. The main structure is illustrated in Figure \ref{fig:structure}. The framework is composed of three core modules: the Temporal Embedding Unit for feature modeling, the Attention Unit for generating snippet-level anomalous attention, and the Multi-branch Supervision Module for learning anomaly completeness and improving localization accuracy. In the subsequent sections, we first give the formulation of the WS-VAD problem and then describe the three modules in detail. Finally, to generate precise anomaly attention, we present an optimization process and provide procedures for how to conduct the training and inference.
\subsection{Problem Formulation}
Following the MIL step, we formulate the WS-VAD problem as follows: let normal videos $V^{n}=\{{v^{n}_{i}}\}^{N}_{i=1}$ and abnormal videos $V^{a}=\{{v^{a}_{i}}\}^{N}_{i=1}$. Each anomaly video is a bag $Y_{a}=1$, containing at least one abnormal instance, while normal videos are marked as $Y_{n}=0$ with only normal instances. The objective of WS-VAD is to learn a function that can assign anomaly scores of snippets $v_{i}$ for each video. To achieve this, we first extract features using pre-trained weights and then perform handling on the extracted features.
In this paper, to ensure consistency with previous methods, we extract snippet-level appearance modality (RGB) features from non-overlapping video volumes containing 16 frames, using the I3D \cite{carreira2017quo} network pre-trained on the Kinetics dataset \cite{kay2017kinetics} as the backbone. The features are 1024-dimensional for each snippet. For $i$-th video with $T$ snippets, we represent the RGB features using matrix tensors $X^{RGB}_{i}\in  R^{T*D}$ (abbr. $X \in R^{T*D}$), where $D$ denotes the dimension of the feature vector.
\subsection{Temporal Embedding Unit}
Anomalies may occur in the short term or over a longer time, therefore both local and global temporal reliance should be considered in WS-VAD tasks. To address this issue, we introduce a temporal encoding unit with two branches: one for capturing local and the other for global dependencies.

Given the feature $ F\in R^{T*D}$, in the global branch, we simply introduce the non-local block proposed in \cite{wang2018non}:
\begin{equation}
	F_{g} = \psi(F),
\end{equation}
where $\psi$ denotes the 1-D non-local operation and $ F_{g}  \in R^{T*\frac{D}{4}} $. As for the local branch, in order to acquire the different time scale local reliance, 1-D convolution operation with dilation (1, 2, 4) is separately used:
\begin{equation}
	F_{l_{1}} =F_{l_{2}} =F_{l_{3}} = \phi(F),
\end{equation}
where $\phi$ denotes the dilated convolution and $F_{l_{i=1}^{3}} \in R^{T*\frac{D}{4}}$.

Then $F_{l}$ and $F_{g}$ are concatenated in feature dimension and get $F^{*}\in R^{T*D}$. A temporal convolution layer is subsequently applied on $F^{*}$ to aggregate features. Finally, with a residual connection, original feature $F$ and $F^{*}$ are simply fused by add operation and acquire enhanced feature $F_{e}$. Due to this temporal embedding unit is also widely used in other WS-VAD methods \cite{tian2021weakly,liu2023weakly,wu2022self}, thus we just briefly state and use it as our baseline in the ablation experiment.
\subsection{Anomalous Attention Unit}
The information surrounding a single snippet is crucial and can help to effectively detect anomalies at a more granular level. To address the problem of intermediate features not being fully utilized in the multiple instance learning (MIL) pipeline, we propose a snippet-level anomalous attention mechanism.

Specifically, after obtaining the enhanced feature, the Temporal Convolution layer $TC$ is first adopted to fully capture channel-wise dependencies and infuse the local context from the neighborhood snippets. Then to avoid some information not being activated in the whole training process and fully utilizing the semantic information, the LeakyRelu activated function $LR$ is introduced for it can generate a negative value. Thus a basic attention unit can be formulated as:
\begin{equation}
	F_{e}^{(l)}= (TC^{(l)}(F_{e}^{(l-1)}) ; LR),
\end{equation}
where $F_{e}^{(l-1)}$ indicates the feature output from the $(l-1)^{th}$ basic unit and the whole attention unit is the stack of this basic unit. The feature dimension of the final TC layer is 1, which means $F \in R^{T*1}$. Then a sigmoid function is used to obtain normalized anomalous attention $A \in R^{T*1}$. This setting enables our method to use the attention normalization term to obtain highly confident snippets.

\subsection{Multi-branch Supervision Module}
\begin{figure}[t]
	\centering
	\includegraphics[width=\linewidth]{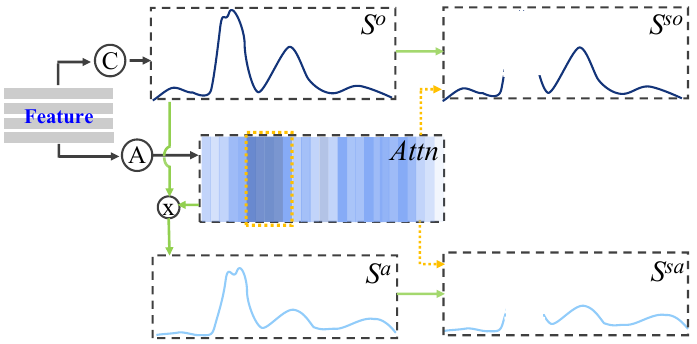}
	\caption{Multi-branch Supervision Module. Features are fed into Classifier \textit{\textbf{C}} and in the meantime directly utilized for snippet-level Anomalous Attention generation \textit{\textbf{A}}. The most discriminative part will be suppressed \textcolor{orange}{(Orange) }and hard abnormal snippets are given more focus.}
	\label{fig:module}
\end{figure}

Multi-Instance Learning is widely known \cite{lv2023unbiased,zhang2022exploiting,li2022self} to suffer from numerous false alarms which are caused by the snippet-level detector to exhibit bias towards abnormal snippets with simple context. Thus intuitively, if reducing the focus on the most discriminative segments, we may effectively explore the completeness of anomalies and challenging snippets. Taking the account of the most discriminative segment, which may include crucial background information of the current video, our initial attempt is to assign lower attention to this discriminative segment and keep contextual information, however, it did not yield satisfactory results (as detailed in the experimental section). Then, we discovered that directly removing the most discriminative segment can satisfactorily enhance the overall detection performance. Thus with the enhanced features $F_{e}$ and anomalous attention $A$, a multi-branch supervision module is designed, shown in Figure \ref{fig:module}. 

\textbf{Original} abnormal scores are directly obtained from the classifier with 3-layer MLP and the nodes are 512, 128, and 1 respectively. Also, each layer is followed by a ReLU and a dropout function. We denote the original anomalous scores as $S^{o}$. Then \textbf{Attention-based} abnormal scores $S^{a}$ can be acquired by element-wise multiplication:
\begin{equation}
	S^{a}= A * S^{o},
\end{equation}
with attention A utilized in $S^{a}$, only anomaly activity is considered and normal events have been suppressed.

Then for the purpose of avoiding the training process dominated by discriminative snippets and better learning the whole video, we calculate the \textbf{Suppressed original} abnormal scores $S^{so}$ and \textbf{Suppressed Attention-based} abnormal scores $S^{sa}$. Concretely:
\begin{equation}
	S^{so}_{ij}=
	\begin{cases}
		S^{o}_{ij},& A_{ij}< \theta_{i}, \\
		0,& otherwise,
	\end{cases}
	\label{eqn:so}
\end{equation}
where $i \in (1,N)$ denotes the $i^{th}$ video and $ j \in (1, T)$ denotes the $j^{th}$ snippets in current sequence. Besides, $\theta$ is a floating value based on the max and min value of the current $A_{i}$ sequence:
\begin{equation}
	\theta_{i}=[max(A_{i})-min(A_{i})]*\epsilon + min(A_{i}),
	\label{eqn:eplilon}
\end{equation}
and $\epsilon$ is a suppressed rate. The reason for processing $\theta$ in this manner is anomaly are diverse and some unobvious anomalies may gain a low anomalous attention. Thus using a fixed parameter as threshold is unreliable. Finally, for abnormal scores $S^{sa}$, it is handled in a similar manner to $S^{so}$:
\begin{equation}
	S^{sa}_{ij}=
	\begin{cases}
		S^{a}_{ij},& A_{ij}< \theta_{i}, \\
		0,& otherwise,
	\end{cases}
\label{eqn:sa}
\end{equation}
where the value of $\theta_{i}$ is equal to its value in Equation (\ref{eqn:eplilon}).

\subsection{Optimizing Process}
\subsubsection{Constraints on Attention} We hope the attention is anomaly-specific, thus the distribution of the anomalous attention should be similar to the final anomalous scores. For negative bags with label $Y_{n}= 0$, the guide loss can be defined as:
\begin{equation}
	L_{guide}^{neg}=\delta(A^{neg},\{0\cdots0\}),
\end{equation}
where $\delta$ means a similarity metric function and we utilize mean square error (MSE). $\{0\cdots0\}$ denotes a sequence that only consists of 0 and has the same size as $A^{neg}$. 

Due to the lack of reliable predictions during the initial training phase, for the positive bag that the label is $Y_{a}= 1$ and contains both normal and abnormal instances, the guide loss is:
\begin{equation}
	L_{guide}^{pos}=
	\begin{cases}
		\delta(A^{pos}, S^{pos}_{o}),& if \; step< M, \\
		\delta(A^{pos},\{0,1,1\cdots1,0\}),& otherwise
	\end{cases}
	\label{eq:posss}
\end{equation}
where $M$ represents the number of training iterations and the sequence $\{0,1,1\cdots1,0\}$ can be obtained by:
\begin{equation}
	1/0=
	\begin{cases}
		1,& if \; S^{pos}_{o}> 0.5, \\
		0,& otherwise,
	\end{cases}
\end{equation}

Besides, due to anomaly is sparse, we also utilize a normalization loss $L_{norm}$ to make the attention more polarized: 
\begin{equation}
	L_{norm}=\Vert A^{pos} \Vert_{1},
\end{equation}
where $\Vert \cdot \Vert_{1}$ is a L1-norm function. 
\subsubsection{Constraints on Video-level Supervision} We apply the widely used binary classification loss on $ S^{o}$, $S^{a}$, $S^{so}$ and $S^{sa}$:
\begin{equation}
	L_{c}= \eta (S, Y),
\end{equation}
where $\eta$ is a binary cross-entropy loss and:
\begin{equation}
	L_{c}^{all}= \alpha(L_{c}^{o}+L_{c}^{a})+(1-\alpha)(L_{c}^{so}+L_{c}^{sa}).
	\label{eqn:alpha}
\end{equation}
\subsection{Network Training and Testing}
\subsubsection{Training.} The same amount of normal and abnormal videos are combined as a batch and fed into our model. Also, we incorporate the temporal smoothness and sparsity constraints term that are also commonly used in other WS-VAD methods and defined as:
\begin{equation}
	\begin{split}
		L_{sm}&= \sum\nolimits^{T-1}_{j=1} (S_{j}-S_{j+1})^2, \\
		L_{sp}&= \sum\nolimits^{T}_{j=1} S_{j},
	\end{split}
\end{equation}
where $L_{sm}$ and $L_{sp}$ are only applied on $ S^{o}$ and $S^{a}$ branches. And the final loss for the training process is:
\begin{equation}
	L= L_{c}^{all} + \gamma L_{norm} + L_{guide}^{neg} + L_{guide}^{pos} + \mu L_{sm} + L_{sp}.
\end{equation}
\subsubsection{Testing.} The testing videos are input into our network and the final predictions are calculated by:
\begin{equation}
	S_{test} = S^{a}.
\end{equation}
Finally, the snippet labels are assigned to the frame level.
\section{Experiments}
\subsection{Datasets and Evaluation Metrics}
\subsubsection{Dataset}
Similar to previous approaches, we assess the performance of our method using two large datasets specifically designed for video anomaly detection: UCF-Crime \cite{sultani2018real} and XD-Violence \cite{wu2020not}.

\textbf{UCF-Crime} dataset is a large-scale dataset of 128 hours of videos. It consists of 1900 long and untrimmed real-world surveillance videos, with 13 realistic anomalies such as fighting, road accident, burglary, robbery, etc. as well as normal activities. The training set includes 800 normal and 810 abnormal videos with video-level labels, while the testing set has 140 normal and 150 abnormal videos, which are annotated at the frame-level.

\textbf{XD-Violence.} The XD-Violence is a more challenge dataset includes a total of 4754 videos collected from both movies and YouTube (in-the-wild scenes) with a variety of scenarios. There are 2405 violent videos and 2349 non-violent videos in the dataset. The training set has 3954 videos while the test set has 800 videos, consisting of 500 violent and 300 non-violent videos.
\begin{table}[t]\centering
	\caption{Comparison of the frame-level performance AUC on UCF-Crime testing set with previous methods. $^{\dagger}$ means we re-train the code with the open-source features.}
	\resizebox{\linewidth}{!}{
		\begin{tabular}{|c|cc|c|c|c|}
		\hline
		& \multicolumn{3}{c|}{Method} & \multicolumn{1}{c|}{Feature (RGB)} & \multicolumn{1}{c|}{AUC (\%)} \\ \hline
		\multirow{2}{*}{Un.} & \multicolumn{2}{c|}{GCL \cite{zaheer2022generative}}  & CVPR 2022   & ResNext \cite{xie2017aggregated}   & 74.20   \\
		&\multicolumn{2}{c|}{S3R*\cite{wu2022self}}                             & ECCV 2022& I3D & 79.58  \\ 
		\hline
		\multirow{15}{*}{Weakly} & \multicolumn{1}{l|}{\multirow{5}{*}{Two-stage}} & GCN \cite{zhong2019graph} & CVPR 2019& TSN \cite{wang2018temporal} & 82.12   \\
		& \multicolumn{1}{l|}{}   & MIST\cite{feng2021mist}                & CVPR 2021  & I3D  & 82.30   \\
		& \multicolumn{1}{l|}{}& MSL \cite{li2022self}& AAAI 2022 & I3D& 85.30  \\
		& \multicolumn{1}{l|}{} & MSL  \cite{li2022self} & AAAI 2022& VideoSwin \cite{liu2022video}& 85.62   \\
		& \multicolumn{1}{l|}{}  & CU-Net \cite{zhang2022exploiting} & CVPR 2023 & I3D & 86.22\\ \cline{2-6}
		& \multicolumn{1}{l|}{\multirow{10}{*}{One-stage}} & Sultani et al. \cite{sultani2018real}  & CVPR 2018 & I3D  & 76.21  \\
		& \multicolumn{1}{l|}{}& CLAWS \cite{zaheer2020claws} & ECCV 2020& C3D \cite{tran2015learning} & 83.03 \\
		& \multicolumn{1}{l|}{}  & RTFM  \cite{tian2021weakly}& ICCV 2021 & I3D& 84.30\\
		& \multicolumn{1}{l|}{} & DAR  \cite{liu2022decouple}            & TIFS 2022 & I3D& 85.18 \\
		& \multicolumn{1}{l|}{}& NG-MIL \cite{park2023normality} & WACV 2023 & I3D & 85.63 \\
		& \multicolumn{1}{l|}{}& S3R \cite{wu2022self}  & ECCV 2022 & I3D & 85.99  \\
		& \multicolumn{1}{l|}{}                           & UR-DMU$^{\dagger}$ \cite{zhou2023dual} & AAAI 2023 & I3D& 86.34 \\
		& \multicolumn{1}{l|}{}& UR-DMU \cite{zhou2023dual} & AAAI 2023 & I3D & \textbf{86.97}  \\
		 \cline{3-6} 
		& \multicolumn{1}{l|}{}  & \multicolumn{2}{c|}{\textbf{Ours (Pytorch)}} & \multicolumn{1}{c|}{I3D}& \multicolumn{1}{c|}{86.19}\\  
		\cline{3-6}  
		& \multicolumn{1}{l|}{}  & \multicolumn{2}{c|}{\textbf{Ours (MindSpore)}} & \multicolumn{1}{c|}{I3D}& \multicolumn{1}{c|}{84.94}\\   
		\hline
	\end{tabular}}
\end{table}

\subsubsection{Evaluation metrics}
Consistent with prior researches, we employ the area under the receiver operating characteristic curve (AUC) to evaluate the effectiveness of our method on the UCF-Crime dataset. Meanwhile, we adopt the precision-recall curve and the corresponding area under the curve (average precision, AP) to evaluate our approach on the XD-Violence dataset. Additionally, in the ablation study, we also report the AUC and AP results on the anomaly subset, following the approach outlined in \cite{wu2021learning,feng2021mist}.

\subsection{Implementation Details}
The whole model can run on a single RTX 2080Ti GPU. During the training process, we utilize the Adam optimizer \cite{kingma2014adam} with a learning rate of 0.0001 and a weight decay of 5e-4. Our batch size is set at 32 with 4000 iterations. In equation (\ref{eq:posss}), M is set to 400. The number of basic units in the Anomalous Attention Unit is 2. And the dropout rate of the MLP part is set as 0.7. We set the hyperparameters $(\epsilon,\alpha,\gamma,\mu)$ to (0.2, 0.8, 0.8, 0.01) in our paper. For $\gamma$ and $\mu$ which are the weights of the loss item, their values are set by a simple balance of the overall loss function without elaborately fine-tuning. The ablation of $\epsilon$ and $\alpha$ will be shown later. In addition to utilizing PyTorch, we also infer our method using Huawei's AI framework, Mindspore \cite{Mindspore}, on the UCF-Crime dataset. 
\subsection{Result on UCF-Crime}
We evaluate the AUC performance of our method on the UCF-Crime dataset. Specifically, RGB features with a 10-crop augmentation are applied which is consistent with the existing approaches. Our method has shown a 1.89\% improvement compared to the one-stage MIL-based method RTFM \cite{tian2021weakly}, and a 0.57\% improvement compared to the two-stage self-training method MSL \cite{li2022self} based on 3D-transformer \cite{liu2022video}. That is, our approach demonstrates superior performance due to its ability to preserve informative snippet-level features and employ a soft attention mechanism, surpassing both regular MIL-based models and even self-training methods. As for a typical UR-DMU \cite{zhou2023dual} method, which introduces two additional memory modules to better store representative patterns, we re-train the code released by the author with the suggested feature (open source and utilized in our method) and get an 86.34\% AUC value. It is important to note that AUC typically demonstrates optimistic results when dealing with class-imbalanced data, such as cases with numerous negative samples. In other words,  in terms of AUC performance, on the UCF-Crime dataset which normal data accounts for a significant proportion of test videos, our method achieves a comparable result, with only a 0.15\% difference. But when it comes to anomaly detection and localization, no matter whether the distribution of anomaly is dispersed or occupies a large portion of the whole video, our method outperforms UR-DMU \cite{zhou2023dual} due to our emphasis on identifying snippet-level abnormalities, and will be demonstrated in the subsequent subsection \ref{sec:Qua}. 
\begin{table}[t]
	\centering
	\caption{Comparison of frame-level AP performance on the XD-Violence validation set.}
	\resizebox{\linewidth}{!}{
		\begin{tabular}{|c|c|c|c|}
			\hline
			\multicolumn{2}{|c|}{Method}& Feature & AP(\%) \\
			\hline
			Sultani et al. \cite{sultani2018real}& CVPR 2018 & RGB&  73.20\\
			HL-Net \cite{wu2020not}&ECCV 2020&    RGB&    73.67 \\
			HL-Net \cite{wu2020not}& ECCV 2020&   RGB+Audio&   78.64\\
			RTFM \cite{tian2021weakly}& ICCV 2021& RGB & 77.81 \\
			MSL \cite{li2022self}&AAAI 2022 &RGB & 78.28 \\
			DAR  \cite{liu2022decouple} &TIFS 2022& RGB &78.94\\
			DAR \cite{liu2022decouple} &TIFS 2022& RGB+Audio &79.32\\
			NG-MIL \cite{park2023normality}&WACV 2023& RGB & 78.51 \\
			S3R  \cite{wu2022self}&ECCV 2022& RGB & 80.26\\
			CU-Net  \cite{zhang2022exploiting}&CVPR 2023& RGB & 78.74 \\
			CU-Net  \cite{zhang2022exploiting}&CVPR 2023& RGB+Audio (concat) & 81.43 \\
			Pang et al.  \cite{pang2021violence} &ICASSP 2021& RGB+Audio &81.69\\
			UR-DMU  \cite{zhou2023dual}&AAAI 2023& RGB & 81.66 \\
			UR-DMU  \cite{zhou2023dual}&AAAI 2023& RGB+Audio  (concat)& 81.77 \\
			\hline
			\multicolumn{2}{|c|}{\textbf{Ours}}& RGB & \textbf{83.59} \\
			\multicolumn{2}{|c|}{\textbf{Ours}} & RGB+Audio (concat)& \textbf{83.77} \\
			\multicolumn{2}{|c|}{\textbf{Ours}} & RGB+Audio (with \textit{TC})& \textbf{84.23} \\
			\hline
	\end{tabular}}
	\label{tab:xd}
\end{table}
\subsection{Result on XD-Violence}
Table \ref{tab:xd} displays the AP scores of state-of-the-art methods on the XD-Violence dataset. The feature we used is in RGB modality from the I3D network with 5-crop augmentation provided by \cite{wu2020not}. Compared to the latest work UR-DMU \cite{zhou2023dual}, which employs additional normal and abnormal memory blocks to acquire more discriminative features, our method achieves a significant improvement of 1.83\%. Furthermore, to test the robustness of our method when using multi-modal features such as RGB and Audio (to be consistent with previous works, we also concatenate them), resulting in an AP score of 83.77\% for our technique which exceeds all the existing works. Additionally, we have implemented a straightforward fusion procedure where the RGB modality is first fed into a temporal convolution (\textit{TC}) module and then concatenated with the original audio feature in the feature dimension. We adopt this approach because the extracted RGB feature is deemed more significant, and there is a domain gap between the final detection task and the extracted network. Introducing a temporal embedding layer aids in aligning the feature to be more task-oriented for the detection task. The new state-of-the-art detection performance of 84.23\% verify our consideration. All these results demonstrate the efficacy of our method in identifying anomalous events, particularly those with disturbances.
\subsection{Ablation Study}
\subsubsection{Impact of the Suppressed manner}
As mentioned earlier, our goal is to suppress the most discriminative parts of the entire video while retaining some contextual information when using the MIL pipeline. However, after conducting experiments, we observed that thoroughly removing discriminative snippets actually resulted in greater improvements. The results are presented in Table \ref{tab:beta}, where we examined different rates of dropped snippets on the XD-Violence dataset. To provide more details, we conducted experiments using $\beta*S^{so}$ or $\beta*S^{sa}$ in place of 0 for the ``otherwise'' case in equations \ref{eqn:so} and \ref{eqn:sa}. We find that as the value of $\beta$ decreases, the performance increases. The experimental results are not consistent with our anticipated outcomes, and we attribute this discrepancy to the background information already better modeled in less discriminative segments.
\begin{table}[t]
	\centering
	\caption{Ablation study of the extent of suppression of discriminative snippets. The smaller $\beta$ is, the greater level of suppression.}
	\label{tab:beta}
	\begin{tabular}{|c|cccccc|}
		\hline
		$\beta$ & 0 & 0.025 & 0.05 & 0.10   & 0.15  & 0.20 \\
		\hline
		XD-Violence & 83.59 & 82.29 &  81.08 & 81.84 & 81.21 & 80.28 \\
		\hline
	\end{tabular}
\end{table}
\begin{table}[t]
	\centering
	\caption{Ablation study on the effectiveness of the different components in the optimizing process on both dataset.}
	\label{tab:ablaLoss}
	\resizebox{\linewidth}{!}{
		\begin{tabular}{|c|cccccc|cc|}
			\hline
			\multirow{2}{*}{}&\multirow{2}{*}{$L_{c}^{o}$} & \multirow{2}{*}{$L_{c}^{a}$} & \multirow{2}{*}{$L_{c}^{so}$} & \multirow{2}{*}{$L_{c}^{sa}$} & \multirow{2}{*}{$L_{guide}$} & \multirow{2}{*}{$L_{norm}$} & \multicolumn{2}{c|}{Result (\%)}\\
			&& & & & && UCF(AUC) & XD(AP) \\
			\hline
			1&\checkmark& -&-&-&-& -&81.96 & 79.28 \\
			2&\checkmark& \checkmark&-&-&-& -&84.34 & 79.60  \\
			3&\checkmark&\checkmark& -& -&-&\checkmark&83.95 & 78.15 \\
			4&\checkmark& \checkmark& -&- & \checkmark&-&84.73 & 80.20  \\
			5&\checkmark&\checkmark& -&-& \checkmark&\checkmark&84.98 & 80.05 \\
			\hline
			6&\checkmark&\checkmark&\checkmark& \checkmark&-&-& 84.45 & 82.76 \\
			7&\checkmark&\checkmark& \checkmark& \checkmark&\checkmark&-& 85.06 & 81.69 \\
			8&\checkmark&\checkmark&\checkmark& \checkmark& -&\checkmark&85.07 & 81.07 \\
			9&\checkmark&\checkmark& -&\checkmark& \checkmark&\checkmark&85.26 & 81.57 \\
			10&\checkmark&\checkmark&\checkmark&- &\checkmark& \checkmark& 84.35 & 80.92 \\
			11&\checkmark&\checkmark&\checkmark&\checkmark& \checkmark&\checkmark&86.19 & 83.59\\
			\hline
	\end{tabular}}
\end{table}
\begin{table}[t]
	\centering
	\caption{Ablation study on the function of the components in multi-branch supervision module. AUC\_sub and AP\_sub are the scores only using abnormal data which can be used to evaluate the anomaly localization performance.}
	\resizebox{\linewidth}{!}{
		\begin{tabular}{|c|c|c|cccc|}
			\hline
			Datasets & \multicolumn{2}{c|}{Setting}      & AUC   & AP    & AUC\_sub & AP\_sub \\
			\hline
			\multirow{5}{*}{XD}  & 1&$S^{o}$         & 92.37 & 79.28 & 81.86    & 81.63   \\
			&2& $S^{o} + S^{a} $      & 92.88 & 79.60  & 81.20     & 81.19   \\
			&3& $S^{o} + S^{a} + S^{so} $   & 93.68 & 80.92 & 81.16    & 81.84   \\
			& 4&$S^{o} + S^{a} + S^{sa}$ & 93.95 & 81.57 & 81.92    & 82.26   \\
			& 5&ALL         & 94.47 & 83.59 & 83.02    & 84.19   \\
			\hline
			\multirow{5}{*}{UCF} &  6&$S^{o}$          & 81.96 & 19.61 & 62.20  & 22.66   \\
			&7& $S^{o} + S^{a} $      & 84.34 & 29.45 & 66.67    & 31.43   \\
			& 8&$S^{o} + S^{a} + S^{so} $   & 84.35 & 29.11 & 64.52    & 30.63   \\
			& 9&$S^{o} + S^{a} + S^{sa}$  & 85.26 & 30.65 & 67.35    & 32.83   \\
			& 10&ALL          & 86.19 & 31.11 & 68.77    & 33.44  \\
			\hline
	\end{tabular}}
	\label{tab:hybrid}
\end{table}
\subsubsection{Effectiveness of the Optimization Items}In this part, we not only conduct an ablation study to examine the effectiveness of the various components in the optimizing process (as shown in Table \ref{tab:ablaLoss}), but we also present the AUC\_{sub} and AP\_{sub} values on the anomaly subset to analyze the function of the Multi-branch Supervision Module (as presented in Table \ref{tab:hybrid}).

With the optimization items of guide and norm combined, in 2 and 5, and lines 6 and 11, the performance has achieved 0.64\% and 1.74\% improvement on the UCF dataset. For XD, guide and norm loss provide 0.45\% and 0.83\% AP gain, indicating the usefulness of these two items and they are more important when the ``suppress'' $S^{so}$ and $S^{sa}$ are utilized. Besides, using the guide and norm separately does not increase the performance significantly or even harm the result (3, 4 and 7, 8), especially in the XD dataset, where the abnormal snippets containing are more evenly distributed.

The introduced attention scores $ S^{a} $ lead to significant performance improvements, particularly on the UCF dataset with gains from 81.96\% to 84.34\% for lines 1 and 2 in Table \ref{tab:ablaLoss}, and improvements in AUC\_sub and AP\_sub from 62.20\% to 66.67\% and 22.66\% to 31.43\% respectively, as shown in line 6 and 7 of Table \ref{tab:hybrid}. Results from the ablation study suggest that detecting abnormalities with BinaryCrossEntropy alone is insufficient. After the attention-based scores are utilized, the UCF-Crime dataset in which the anomaly is more dispersed can obtain a better improvement. But for XD, because WS-VAD is only a two-class problem, the gain is not obvious. 

Besides, performance improvement also occurs on both datasets when the attention-based suppressed abnormal branch $S^{sa}$ is introduced, line 9 of Table \ref{tab:ablaLoss}, items 4 and 9 of Table \ref{tab:hybrid}. The phenomenon is predictable that suppressing the most discriminative snippets would aid the learning of hard snippets, thereby improving the result of anomaly detection. But when the original suppressed scores $S^{so}$ are applied individually, the performance will be harmed, line 10 of Table \ref{tab:ablaLoss}, items 3 and 8 of Table \ref{tab:hybrid}, due to overly severe penalties for the snippets which have low anomalous attention. Finally, the combination of all supervisions leads to our method achieving state-of-the-art performance.
\begin{figure*}[t]
	\centering
	\includegraphics[width=0.92\linewidth]{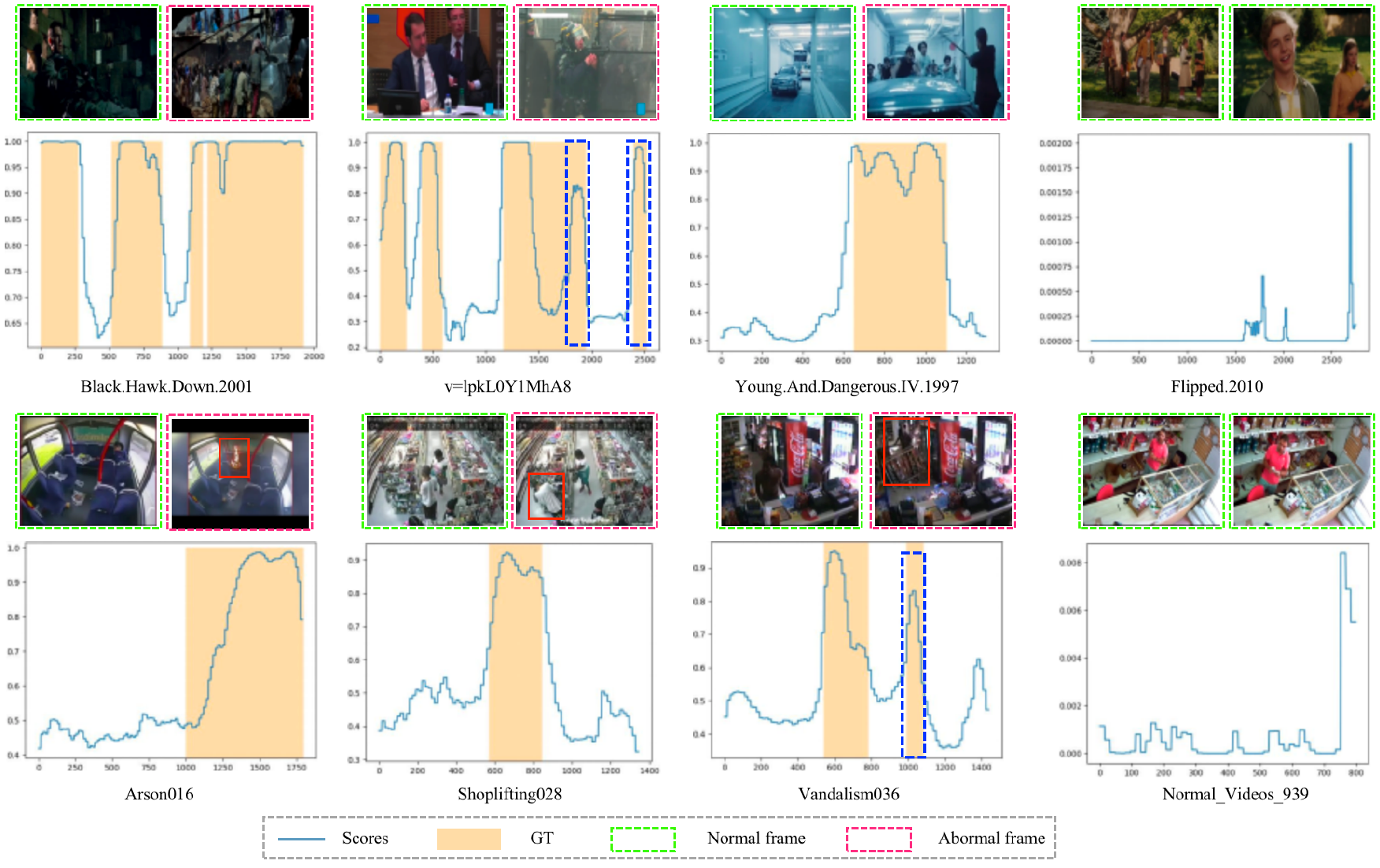}
	\caption{Qualitative results of anomaly detection performance on XD-Violence (Two rows above) and UCF-Crime dataset (Two rows below). }
	\label{fig:Qua}
\end{figure*}
\begin{table}[t]
	\centering
	\caption{Ablation study on the performance that use different values of scores during testing.}
	\resizebox{\linewidth}{!}{
		\begin{tabular}{|c|cccc|cccc|}
			\hline
			\multirow{2}{*}{Datasets} & \multicolumn{4}{c|}{UCF}            & \multicolumn{4}{c|}{XD}             \\
			\cline{2-9}
			& AUC   & AP    & AUC\_sub & AP\_sub & AUC   & AP    & AUC\_sub & AP\_sub \\
			\hline
			W/O  & 85.03 & 29.34 & 66.58    & 31.20   & 94.08 & 81.84 & 82.10   & 82.65   \\
			With & 86.19 & 31.11 & 68.77    & 33.44   & 94.47 & 83.59 & 83.02    & 84.19  \\
			\hline
	\end{tabular}}
	\label{tab:test2}
\end{table}
\begin{figure}[t]
	\centering
	\includegraphics[width=\linewidth]{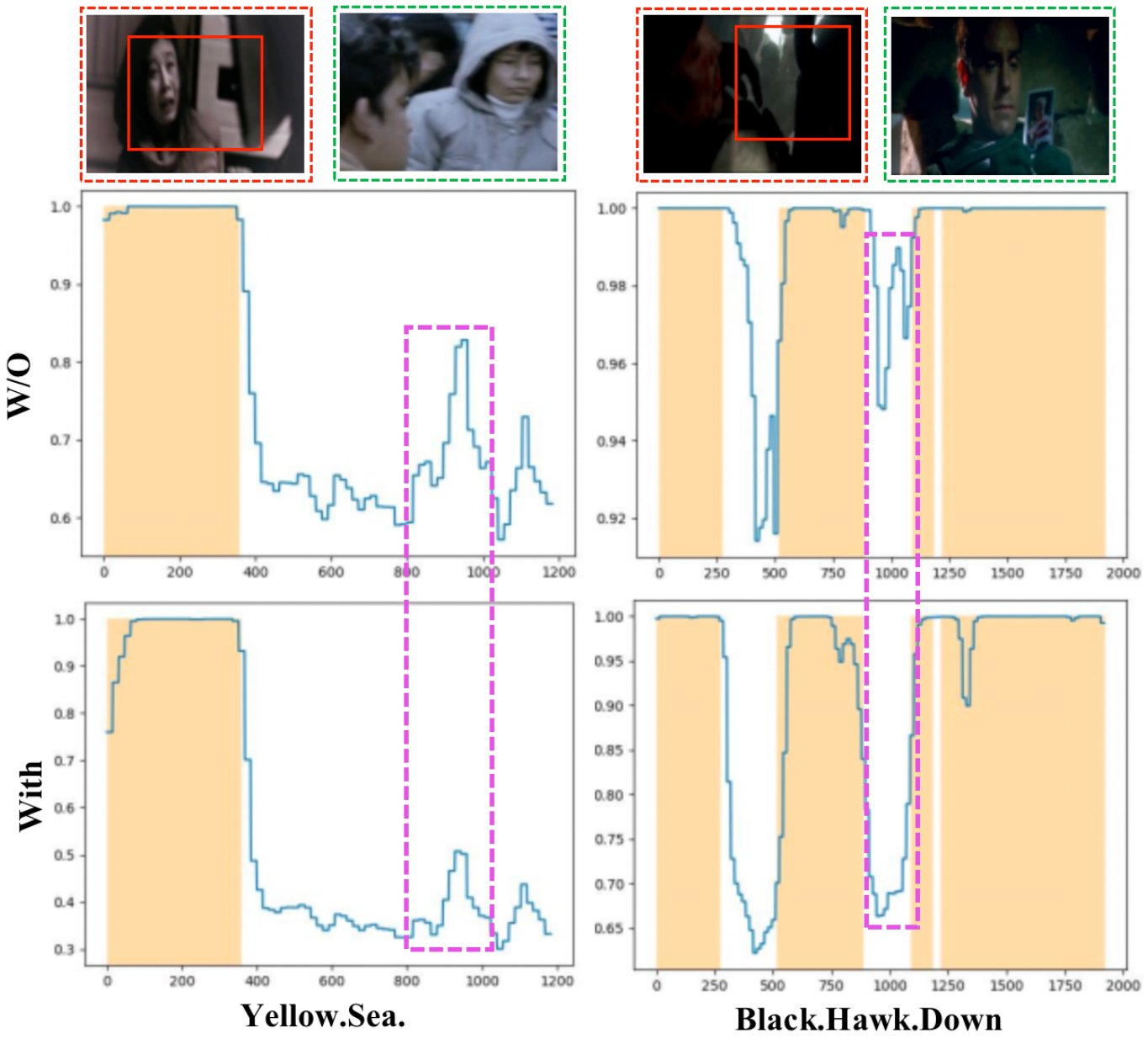}
	\caption{Visualization of obtained curves when using different values of anomaly scores (W/O refers to $S^{o}$ and With refers to $S^{a}$) during the testing phase.}
	\label{fig:test}
\end{figure}
\subsubsection{Effectiveness of the Attention Branch}
In Table \ref{tab:test2}, we compare the results of different scores used in the testing process. Our results show that attention greatly improves performance, no matter in the whole testing sets or only anomaly subsets. Furthermore, we demonstrate the effectiveness of the snippet-level anomaly-specific attention in Figure \ref{fig:test}, where we compare testing video scores with $S^{a}$ and without attention $S^{o}$. Our results indicate that the separate attention module allows the network to identify abnormal frames more accurately, resulting in a more polarized score curve, and can suppress false positives, shown in the \textcolor{violet}{violet} rectangle of Figure \ref{fig:test}. In summary, an attention branch is an effective tool for improving detection performance in differentiating between abnormal and normal classes from the snippet level.
\begin{table*}[t]
	\centering
	\caption{Compasion with UR-DMU under varying numbers of divided segments.}	\label{seg}
	\resizebox{\linewidth}{!}{
		\begin{tabular}{|c|c|cccc|cccc|cccc|cccc|cccc|cccc|}
			\hline
			\multicolumn{2}{|c|}{Segments}      & \multicolumn{4}{c|}{32}           & \multicolumn{4}{c|}{64}           & \multicolumn{4}{c|}{100}          & \multicolumn{4}{c|}{200}          & \multicolumn{4}{c|}{320}          & \multicolumn{4}{|c|}{Avg}            \\
			\hline
			Datasets            & Setting     & AUC   & AP    & AUC\_sub & AP\_sub & AUC   & AP    & AUC\_sub & AP\_sub & AUC   & AP    & AUC\_sub & AP\_sub & AUC   & AP    & AUC\_sub & AP\_sub & AUC   & AP    & AUC\_sub & AP\_sub & AUC    & AP     & AUC\_sub & AP\_sub \\
			\hline
			\multirow{3}{*}{XD} & UR-DMU      & 92.73 & 76.99 & 79.22   & 78.36  & 90.55 & 78.55 & 79.87   & 80.92  & 92.99 & 78.00    & 80.87   & 79.64  & 94.02 & 81.66 & 82.36   & 82.85  & 93.53 & 80.86 & 80.78   & 81.73  & 92.76 & 79.21 & 80.62   & 80.70   \\
			& OURS        & 92.27 & 78.91 & 80.64   & 81.29  & 93.71 & 80.68 & 82.14   & 82.02  & 94.47 & 83.59 & 83.02   & 84.19  & 94.05 & 83.31 & 81.44   & 83.83  & 94.35 & 83.17 & 82.50    & 83.75  & 93.77  & 81.93 & 81.95  & 83.02 \\
			& Margin & 0.46 $\downarrow$& 1.92$\uparrow$  & 1.42  $\uparrow$  & 2.93$\uparrow$   & 3.16  $\uparrow$& 2.13$\uparrow$ & 2.27 $\uparrow$   & 1.10$\uparrow$    & 1.48 $\uparrow$ & 5.59  $\uparrow$& 2.15 $\uparrow$   & 4.55 $\uparrow$  & 0.03$\uparrow$  & 1.65  $\uparrow$& 0.92$\downarrow$   & 0.98$\uparrow$   & 0.82  $\uparrow$& 2.31  $\uparrow$& 1.72$\uparrow$    & 2.02$\uparrow$   & 1.01$\uparrow$  & 2.72$\uparrow$   & 1.33$\uparrow$  & 2.32$\uparrow$\\
			\hline
	\end{tabular}}
\end{table*}
\begin{table}[t]
	\centering
	\caption{Performance of the number of divided snippets during the training stage.}
	\begin{tabular}{|c|cccccc|}
		\hline
		\multirow{2}{*}{Dataset} & \multicolumn{6}{c|}{Segments}\\
		\cline{2-7}
		& 32    & 64    & 100   & 200   & 320   & 400   \\
		\hline
		UCF& 85.24 & 84.93 & 84.40  & 84.75 & 86.19& 85.49 \\
		XD& 78.91 & 80.68 & 83.59& 83.31 & 83.17 & 81.53\\
		\hline
	\end{tabular}
	\label{tab:seg}
\end{table}
\begin{table}[t]
	\centering
	\caption{Comparison with other methods at T = 32 on XD-Violence dataset.}	\label{32}
	\resizebox{\linewidth}{!}{
		\begin{tabular}{|c|ccccc|}
			\hline
			\multicolumn{6}{|c|}{XD-Violence}         \\
			\hline
			\multirow{2}{*}{Result (\%)} & RTFM\cite{tian2021weakly}  & MSL \cite{li2022self}& NG-MIL \cite{park2023normality}& CU-Net \cite{zhang2022exploiting}& OURS  \\
			\cline{2-6}
			& 77.81 & 78.28 & 78.51  & 78.74  & 78.91\\
			\hline
	\end{tabular}}
\end{table}
\begin{table}[t]
	\centering
	\caption{Impact of the hyperparameter $\epsilon$ in equation (\ref{eqn:eplilon}).}\label{epsilon}
	\resizebox{\linewidth}{!}{ 
		\begin{tabular}{|c|ccccccccccc|}
			\hline
			$\epsilon$  & 0.0 &0.05& 0.10 &0.15  & 0.20 &0.25  & 0.30  &0.35  & 0.40&0.45  & 0.50 \\
			\hline
			UCF& 85.45 & 84.44 &86.01 & 85.46 & 86.19 & 85.37 & 85.22 & 85.38 & 85.94 & 85.49 & 84.78\\
			XD& 79.92 & 81.82 & 82.13 & 82.07 & 83.59 & 82.24 & 82.35 & 82.35 & 82.30 & 82.63 & 82.60 \\
			\hline
	\end{tabular}}
\end{table}
\subsubsection{Impact of the number of divided snippets}
In the WS-VAD task, each video is typically divided into non-overlapping snippets during the training stage. In previous researches, T = 32 is set in \cite{tian2021weakly,park2023normality} and in \cite{cao2022adaptive} is 150, and in \cite{zhou2023dual} is set with 200.  Our approach takes snippet-level semantic information into consideration, which suggests that the number of snippets used in the training stage may impact performance. Therefore, we conduct an ablation experiment under different numbers of snippets. Table \ref{tab:seg} demonstrates that the best results are obtained on the XD-Violence dataset when T = 100 with an AP of 83.59, while for the UCF-Crime dataset, the best performance 86.19 is achieved when T = 320.
\begin{figure}[b]
	\centering
	\includegraphics[width=0.80\linewidth]{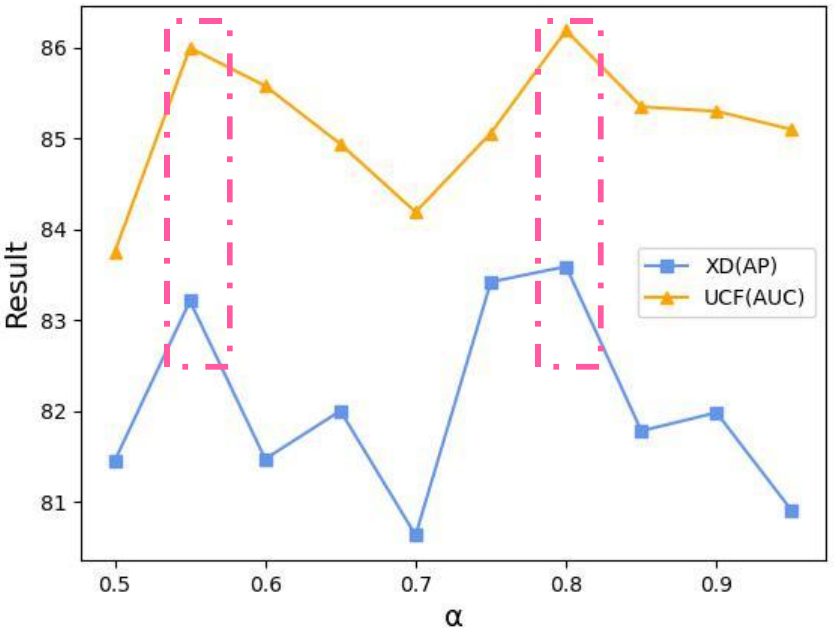}
	\caption{Impact of the hyperparameter $\alpha$ in the total loss function.}
	\label{fig:alpha}
\end{figure}
Furthermore, to ensure fairness in the comparison and provide substantial evidence of the effectiveness of our method, we reproduce the UR-DMU \cite{zhou2023dual} method, which is openly available, under varying settings for the number of divided snippets on the XD-Violence dataset. The presented results, as shown in Table \ref{seg}, establish that our method consistently outperforms UR-DMU. Furthermore, we performed a comparison of our approach against other studies while maintaining the same number of segmented snippets (i.e., T=32). The results are shown in Table \ref{32}, and our method continues to demonstrate superior performance.
\begin{figure*}[t]
	\centering
	\includegraphics[width=\linewidth]{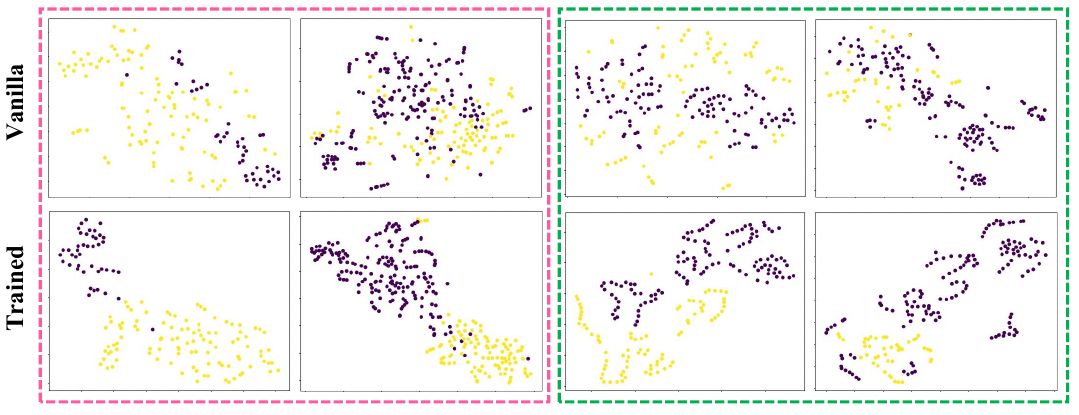}
	\caption{Visualizations of the vanilla features and the output of our model on testing videos (\textcolor{pink}{XD-Violence} and \textcolor{green}{UCF-Crime}).}
	\label{fig:tsne}
\end{figure*}
\begin{figure*}[t]
	\centering
	\includegraphics[width=\linewidth]{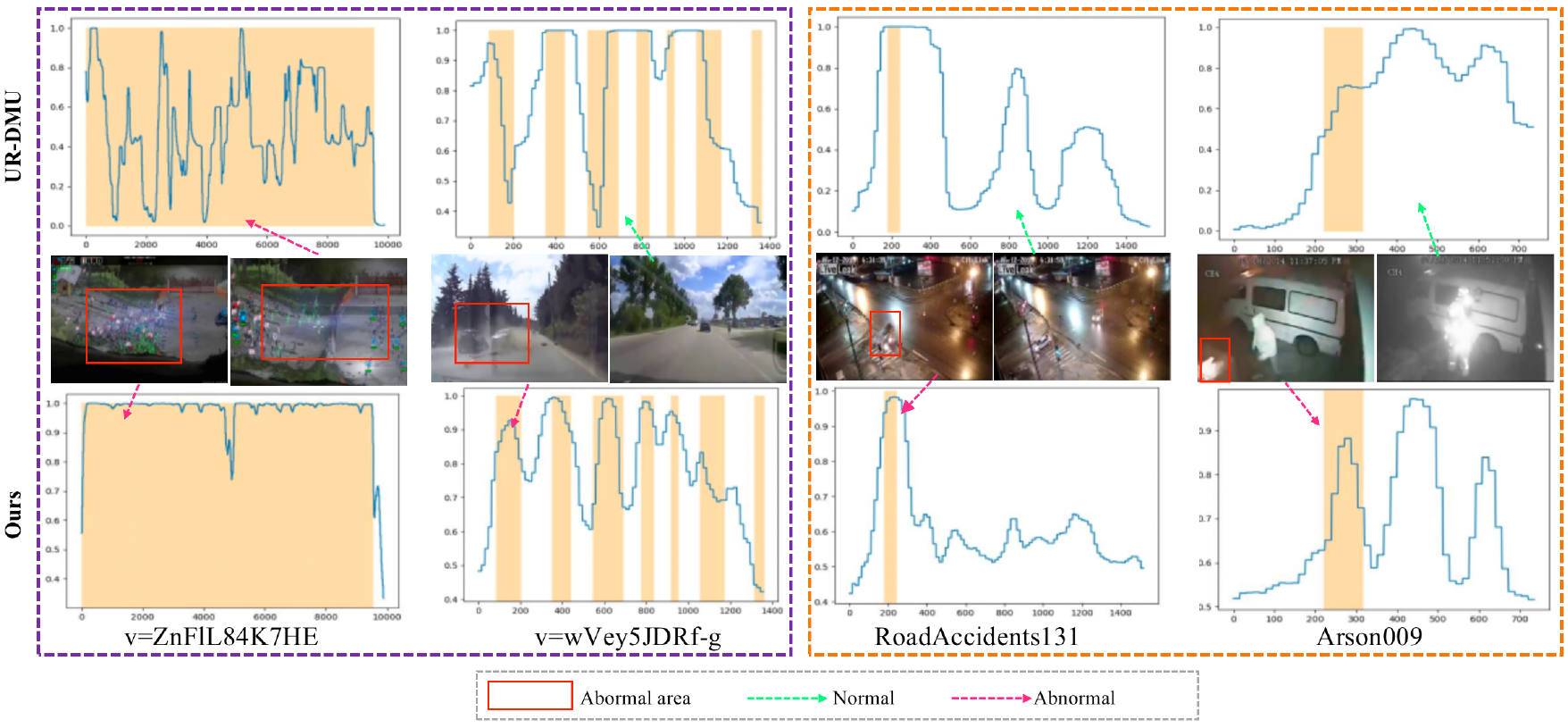}
	\caption{Qualitative results of anomaly detection and localization performance on \textcolor{violet}{XD-Violence} and \textcolor{orange}{UCF-Crime} when compare with UR-DMU \cite{zhou2023dual} method.}
	\label{fig:Qua2}
\end{figure*}
\subsubsection{Effect of iterations M in abnormal guide loss}
The accuracy of abnormal video predictions during the initial training stage is unreliable. Therefore, we employ a soft approach to processing the guide loss for abnormal videos when the iteration is less than $M$. Once the iteration count exceeds $M$, we utilize hard labels with values of 1 and 0 for guiding backpropagation. The effect of parameter $M$ on the final results of both datasets is presented in Table \ref{tab:iters}. Our method shows a rapid convergence, thus we only set the maximum value of the ablation as 700. It is evident that using hard labels to monitor at the beginning will result in a decrease in performance. And when the best parameter settings are reached, subsequent loss settings will also damage performance.
\begin{table}[t]
	\centering
	\caption{Performance of the \textit{M} iterations in equation \ref{eq:posss}.}
	\resizebox{\linewidth}{!}{
		\begin{tabular}{|c|cccccccc|}
			\hline
			\multirow{2}{*}{Dataset} & \multicolumn{8}{c|}{Iteration}  \\
			\cline{2-9}
			& 0     & 100   & 200   & 300   & 400   & 500   & 600   & 700   \\
			\hline
			UCF  & 84.66 & 84.49 & 85.01 & 84.68 &86.19& 84.94 & 85.42 & 85.98 \\
			XD & 81.82 & 81.99 & 81.75 & 82.16 & 83.59 & 82.24 & 81.86 & 81.46\\
			\hline
	\end{tabular}}
	\label{tab:iters}
\end{table}
\subsubsection{Effect of hyperparameter $\alpha$ in the suppressed branch}The hyperparameter $\alpha$ in equation (\ref{eqn:alpha}) regulates the proportion of suppressed information during the optimization process. It also reflects the degree of emphasis placed on challenging anomalies. Ablation of $\alpha$ is shown in Figure \ref{fig:alpha}. It is evident that excessively focusing on non-discriminative components during the optimization procedure will result in a decline in performance. Nevertheless, an interesting improvement in both the XD and UCF datasets was observed when alpha was set to 0.55, \textcolor{pink}{pink} area in the figure.

\subsubsection{Effect of hyperparameter $\epsilon$ in the suppressed branch}The hyperparameter $\epsilon$ in equation (\ref{eqn:eplilon}) determines which segments of the video will be identified as discriminative. Firstly, we would like to state that a smaller value of Ellison indicates a threshold closer to the minimum value of the entire video attention. This means that more video segments are considered discriminative. In the XD dataset, we observed that setting a lower value for Ellison leads to lower final detection results. Once the suppression limit is reached, there is no significant fluctuation in performance. We believe that this situation arises because the number of discriminative segments in a video is limited, and they are truly easy to be recognized. Therefore, as the threshold increases, the actual variation in the recognized discriminative segments is not significant.
\subsection{Qualitative Results}
\label{sec:Qua}
To further assess the effectiveness of our approach, we present qualitative results on two datasets in Figure \ref{fig:Qua}. Eight videos are shown with frame-level anomaly predictions, where the six videos on the left are positive, and the two on the far right are normal. The orange area indicates the ground truth, while cyan lines represent our predictions. Additionally, we highlight normal and abnormal frames in the videos with green and red boxes, respectively. The figure demonstrates that our method has effectively achieved good performance in terms of anomaly detection and event localization. Notably, our method can also effectively detect short-duration abnormal snippets, the blue rectangle shown in the figure. 

We also present t-SNE \cite{van2008visualizing} visualizations depicting the feature distributions on both benchmark test sets. Figure \ref{fig:tsne} displays the results, where abnormal segments are represented by yellow dots and normal features are represented by purple dots. It is evident that the normal and abnormal features are distinctly clustered, and the distance between unrelated features is widened after the training process. This observation demonstrates that with the assistance of our proposed network, instances are effectively differentiated, thereby further validating the efficacy of our framework.
\begin{figure}[t]
	\centering
	\includegraphics[width=\linewidth]{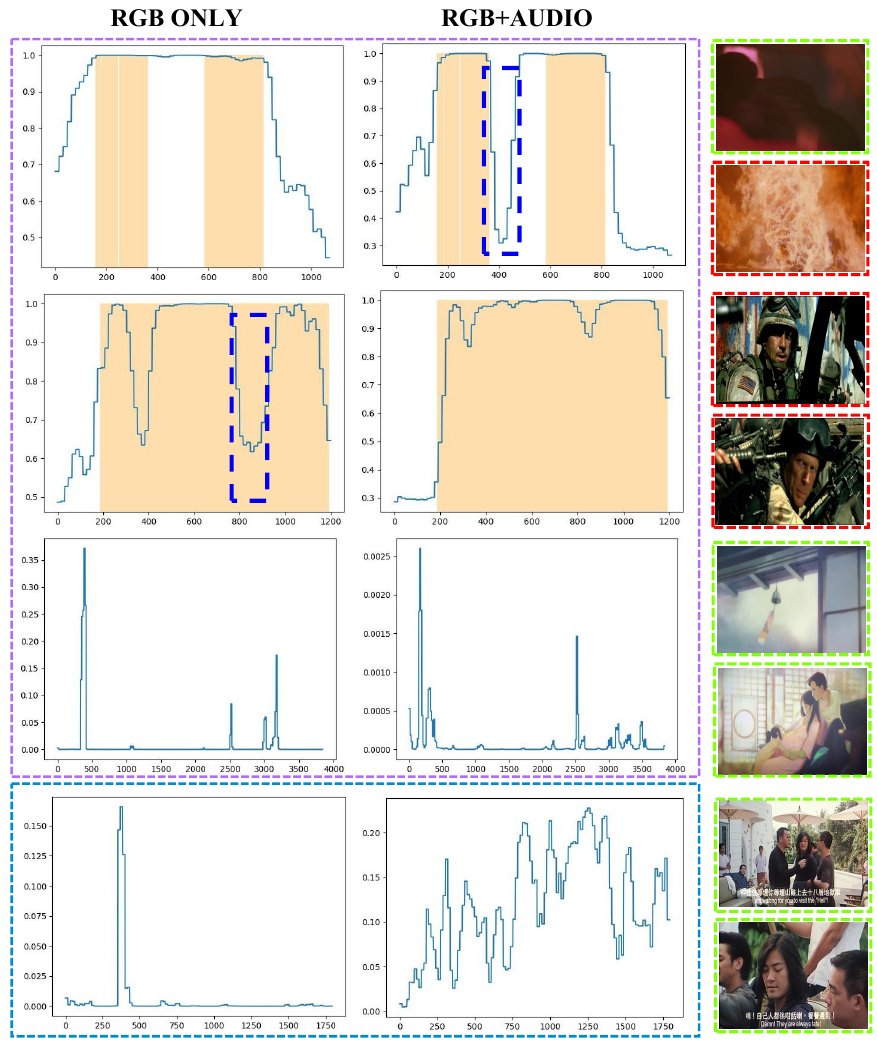}
	\caption{Comparison of anomaly detection results on the XD-Violence dataset using either RGB-only features or RGB-Audio fusion features.}
	\label{fig:fusion}
\end{figure}
Further, we compare our method with UR-DMU \cite{zhou2023dual} on two datasets from the perspective of anomaly detection and localization. The result of UR-DMU is drawn through the best checkpoint released by the author, shown in Figure \ref{fig:Qua2}. Our method demonstrates greater robustness to noise from object changing and scene transforming (even if computer game scenes). When anomalies account for a large proportion of the whole video, our method effectively integrates local information surrounding the anomaly while concurrently restraining the influence of discriminative segments, resulting in a relatively smooth anomaly score curve with small fluctuations (column 1). Additionally, owing to the snippet-level focus, our approach demonstrates improved accuracy in localizing anomalies, particularly when the abnormal snippets within a video are short or the anomaly distribution is dispersed. This is evident from the result presented in columns 2 and 3. In column 4, the ``Arson'' event actually is of short duration, however, the camera videos are collected at night, thus the firefighters wearing reflective vests also are recognized as arson event. Although neither method effectively distinguishes normal snippets, our approach provides a more polarized result and proves that our method can better detect anomalies.

Finally, we present anomaly detection results utilizing various modalities, as depicted in Figure \ref{fig:fusion}. Normal frames are represented by the \textcolor{green}{green} box, while abnormal frames are indicated by the \textcolor{red}{red} box. The fusion of modalities, represented by the \textcolor{violet}{violet} box, provides an enhanced final detection outcome. Conversely, the result represented by the \textcolor{cyan}{cyan} color is detrimental. In the first row, the introduction of audio features enables accurate localization of abnormal events, despite flame elements seemly existing in the \textcolor{blue}{blue} portion of the video clip, with the background sound suggesting that it is actually driving. In the second row, even though there are no obvious cues such as guns or other visual information, the abnormal frame is still detected, due to the presence of gunfire sounds throughout the entire video. The subsequent two rows display normal videos, with the last row revealing an increase in abnormal scores, which may be attributed to the presence of noisy background sound. To sum up, the visualization showcases the benefits of combining visual and acoustic information, highlighting the importance of using a reasonable fusion approach to avoid introducing noise that could potentially compromise the overall performance.
\section{Further consideration}
During the experimental process, we discovered two limitations in our methods. Firstly, the MSE loss function poses an unsatisfactory backpropagation when the attention value approaches 0.5. However, 0.5 is a crucial threshold for distinguishing between normal and abnormal events. Secondly, we explored other new forms of attention modules, such as a fusion of saliency and contextual information, which initially showed promising results during training but ultimately yielded dissatisfactory outcomes. In essence, these situations highlight the need for further improvement in the optimization process designed by us.

\section{Conclusion}
In this paper, we propose a method that takes the snippet-level encoded features into consideration. Concretely, after modeling the original features at global and local levels, an attention mechanism is introduced. Then together with the snippet anomalous attention, a multi-branch supervision module is proposed, in which not only the general predicted scores but also attention-based predictions are utilized. Besides, we also suppress the most discriminative snippets, so the hard portion of the video can be learned and then explored the anomaly completeness. Finally, for better generating anomalous attention, an optimizing process that contains norm and guide items is given. With the combination of components mentioned above, our method achieves state-of-the-art performance on two large benchmark datasets.

\end{document}